%% file: root.tex
\documentclass[letterpaper, 10pt, journal]{ieeetran}                                  

\usepackage{amsmath,amsfonts,amssymb}
\usepackage{algorithmic}
\usepackage{algorithm}
\usepackage{array}
\usepackage[caption=false,font=normalsize,labelfont=sf,textfont=sf]{subfig}
\usepackage{textcomp}
\usepackage{stfloats}
\usepackage{url}
\usepackage{verbatim}
\usepackage{graphicx}
\usepackage{cite}
\usepackage[table]{xcolor}
\usepackage[T1]{fontenc}
\usepackage{microtype}
\usepackage[colorlinks=true,linkcolor=blue,anchorcolor=blue,citecolor=blue,urlcolor=blue]{hyperref}
\usepackage{cleveref}
\usepackage{orcidlink}

\definecolor{best}{RGB}{163, 215, 255}  
\definecolor{secondbest}{RGB}{227, 243, 255} 
\newcommand{\highlight}[2]{\colorbox{#1}{#2}}
\newcommand{\etal}{\textit{et al.}}

\begin{document}

\title{LoC-IN: Real-Time Localization of Cameras \\in Neural Rendered Prior LiDAR Maps}

\author{Joni Vanherck\orcidlink{0009-0008-1312-0382}, Steven Moonen\orcidlink{0000-0002-0935-701X}, Brent Zoomers\orcidlink{0009-0000-7071-8534}, Kobe Werner\orcidlink{0009-0001-2970-228X},\\ Jeroen Put\orcidlink{0000-0003-1122-6366}, Lode Jorissen\orcidlink{0000-0001-7228-9315} and Nick Michiels\orcidlink{0000-0002-7047-5867}%
\thanks{This research was funded by the Special Research Fund (BOF) of Hasselt University (R-14360, R-14436) and FWO fellowship grants (1S80926N, 1SHDZ24N), the Flanders Make's ALARMM SBO project (R-15482), and the Flanders Make's XRTWin SBO project (R-12528). This work was made possible with the support of MAXVR-INFRA, a scalable and flexible infrastructure that facilitates the transition to digital-physical work environments.}%
\thanks{All authors are with the Digital Future Lab, Flanders Make, Hasselt University, 3590 Diepenbeek, Belgium, {\tt\footnotesize \{firstname\}.\{lastname\}@uhasselt.be}}%
}

\maketitle

\input{sec/0_abstract}
\input{sec/1_intro}
\input{sec/2_related}
\input{sec/3_method}
\input{sec/4_eval}
\input{sec/5_conclusion}

\bibliographystyle{IEEEtran}
\bibliography{bib/IEEEabrv, bib/main}

\end{document}

%% file: sec/0_abstract.tex
\begin{abstract}
Accurate camera localization is crucial for robotics, enabling reliable navigation.
Existing visual methods often suffer from drift and scale ambiguity and depend on fiducials or loop closure.
This work introduces a real-time method for localizing a camera within a pre-captured, highly accurate colored LiDAR point cloud.
We bypass drift by using a recent neural rendering method and establishing 2D-3D correspondences between these renders and the live camera frames, achieving accurate camera-to-LiDAR registration. 
The neural rendering technique narrows the domain gap between synthetic and real images, reducing occlusion and background artifacts to improve feature matching.
Due to the LiDAR prior, tracking is performed with the correct metric scale in a global coordinate system without a warm-up period. 
Two real-time variants are presented: Render\&Match and Prebuild\&Localize.
The latter one integrates with existing SLAM frameworks and does not rely on GPU computation during tracking.
We demonstrate improved results on ScanNet++ and the Oxford Spires dataset. 
\end{abstract}

%% file: sec/1_intro.tex
\section{Introduction}
\label{sec:intro}

Camera localization, the process of determining a camera's precise position and orientation within an environment, is essential to both robotics and Extended Reality (XR), enabling autonomous navigation, immersive digital-twin alignment, and applications like robotic teleoperation.

However, accumulating drift remains a severe problem in modern tracking techniques, even when operating in confined spaces. 
For example, state-of-the-art multi-camera IMU headsets still suffer drift of several centimeters across small spaces~\cite{singh_evaluation_2025}. This error is highly problematic because even minor discrepancies can cause systems to miss targets or collide with obstacles. Even worse, this drift accumulates over time, continuously increasing the error until complete divergence. 
To mitigate drift, traditional methods often rely on intrusive artificial landmarks (e.g., fiducial markers~\cite{garrido-jurado_automatic_2014}) that are impractical or not allowed in industrial environments.
Alternatively, frameworks employ loop closing~\cite{mur-artal_fast_2014} to correct accumulated errors upon revisiting known areas. However, constructing a well-covered, loop-closed SLAM map online takes time, causes abrupt pose jumps during optimization, and fails in low-overlap areas like long hallways.

To bypass the vulnerabilities of online map construction, prior scene reconstructions have been exploited for the localization task~\cite{zhang_visual_2026,panek_meshloc_2022,sarlin2019coarse,hyeon2021pose, munoz-salinas_lidar_2025}. However, a fundamental limitation of these reconstructions is that they do not provide millimeter-accurate maps to start with~\cite{bai_colmap-pcd_2024}. We instead leverage a prior 3D scan captured with a static LiDAR scanner. These scanners produce sub-millimeter-accurate, colorized point clouds with immediate metric scale, providing a strong starting point for camera localization. While these devices are expensive, only a single scan composed of multiple concatenated scans is required, making one-time rental or outsourcing feasible. Because these maps consist of unorganized, floating points, they cannot be directly projected into coherent images for camera tracking~\cite{vanherck_real-time_2025}. Consequently, a domain gap exists between the live camera images and the 3D point cloud data. Prior work has attempted to resolve this by direct cross-modal matching~\cite{cattaneo_cmrnet_2019,cattaneo_cmrnext_2025,feng_2d3d-matchnet_2019,li_2d3d-matr_2023}. These methods fail to generalize to unseen scenes, or introduce computational overhead, making them too inefficient to run in real-time.  Others have tried to resolve this by converting the point clouds into surface meshes~\cite{zhang_visual_2026,panek_meshloc_2022}, or optimizing radiance fields of the scene~\cite{zhang_visual_2026, munoz-salinas_lidar_2025}. However, meshing algorithms inherently degrade the geometric accuracy of the original scan, whereas radiance fields require a collection of posed images with strong coverage, which is typically unavailable. 

To address these limitations, we propose a real-time camera localization system operating within pre-captured, high-accuracy, colored LiDAR point clouds. Our pipeline frames camera-to-LiDAR registration as a 2D-2D matching problem by rendering synthetic views of the LiDAR scan, using a neural rendering module~\cite{pointersect,vanherck_real-time_2025,dai2020neural} to bridge the domain gap caused by occlusions and sparsity. Concretely, our work makes three main contributions:
\textbf{(i)} We present a real-time pipeline for camera-to-LiDAR registration by matching live images to neural-rendered views, yielding drift-bounded, globally registered localization from the very first frame. 
\textbf{(ii)} We demonstrate that lightweight, classic features like ORB~\cite{rublee_orb_2011} and XFeat~\cite{Potje_2024_CVPR} are sufficient for robust tracking within a prior LiDAR map, avoiding slow, resource-intensive neural matching networks.
\textbf{(iii)} Building on this pipeline, we propose two distinct localization methods: \textbf{Render\&Match (R\&M)}, an online pipeline that directly matches live frames with dynamically rendered views to estimate camera poses, and \textbf{Prebuild\&Localize (P\&L)}, a preprocessing-based variant that pre-computes a sparse feature map from the neural-rendered point cloud, allowing to efficiently track the camera pose with the localization module of traditional, unmodified SLAM frameworks, without requiring active GPU-accelerated neural rendering during tracking.

%% file: sec/2_related.tex
\section{Related work}
\label{sec:related_work}

\textbf{Tracking-Based solutions \& Multi-Modal SLAM} --
Early approaches registered a moving camera to a pre-captured LiDAR map by first generating a local map via Visual Odometry (VO) or Visual-Inertial Odometry (VIO) and subsequently aligning it with the global 3D geometry~\cite{caselitz_monocular_2016, kim_stereo_2018, yu_monocular_2020}. However, these frameworks are highly sensitive to drift accumulation within the local tracking module, and cannot recover if tracking is lost. 
To bypass these failure modes, simultaneous online multi-modal SLAM systems~\cite{zheng_fast-livo2_2025, wisth_vilens_2023} combine visual, inertial, and LiDAR streams concurrently. While highly accurate, these online solutions require heavy onboard LiDAR sensors, depend on accurate cross-sensor extrinsic calibration and time synchronization, making them impractical for lightweight, mobile systems. 

\textbf{Prior Map Representations} --
When a prior map of the environment is available, localization can be framed as an image-retrieval task followed by fine pose refinement. Coarse-to-fine pipelines like HLoc~\cite{sarlin2019coarse} utilize robust global image descriptors~\cite{arandjelovic_netvlad_nodate, berton_eigenplaces_2023, berton_megaloc_nodate} to identify database images from Structure-from-Motion (SfM) reconstructions. For finer pose optimization, recent frameworks construct alternative 3D scene representations such as explicit surface meshes~\cite{panek_meshloc_2022,zhang_visual_2026}, point clouds~\cite{hyeon2021pose,zhang_visual_2026}, Neural Radiance Fields~\cite{zhang_visual_2026}, or 3D Gaussian Splatting~\cite{khatib_gsvisloc_2025}, matching frames against synthesized environments. 
However, explicit surface meshing degrades geometric fidelity, while, other representations typically rely on scale-ambiguous SfM reconstructions, which demand massive preprocessing overhead, and fail to guarantee centimeter-level global accuracy. Additionally, these methods operate on single images only without utilizing temporal information that is available in typical sequences, which is why we employ a tracking phase.

\textbf{Render-and-Match Frameworks} --
To retain the native geometric fidelity of a 3D scan, render-and-match techniques project point data into 2D virtual views to convert 3D-to-2D registration into an intra-modal 2D–2D feature matching task. For offline processing, pipelines like ScanNet++~\cite{yeshwanthliu2023scannetpp} integrate raw RGB laser scans directly into the COLMAP~\cite{schoenberger2016sfm} bundle adjustment pipeline, leveraging dense photometric errors for pose refinement. 
Other alternatives attempt to project colored point clouds onto the camera frame to establish tracking~\cite{munoz-salinas_lidar_2025, michiels_tracking_2024}. Because raw LiDAR point clouds consist of disconnected, floating points, their direct projection yields renders with holes and background leakage. These methods require a known initial pose and a rough trajectory, or use artificial landmarks. In contrast, our pipeline integrates real-time neural point cloud rendering to create hole-filled renders, enabling robust matching without markers or tracking priors. 
We compare our method to the closest related works~\cite{munoz-salinas_lidar_2025,zhang_visual_2026} in \cref{tab:related}.

\begin{table}[h]
    \centering
    \caption{Comparison to the closest related works\cite{munoz-salinas_lidar_2025, zhang_visual_2026}.}
    \begin{tabular}{l|ccc}
    \hline
     & \cite{zhang_visual_2026} & \cite{munoz-salinas_lidar_2025} & Ours \\
    \hline
    Real-time processing & -- & \checkmark & \checkmark \\
    Monocular data only & \checkmark & -- & \checkmark \\
    Tracking module & -- & \checkmark &  \checkmark\\
    No rough trajectory/initialization required & \checkmark & -- & \checkmark \\
    Neural rendering without scene-specific training & -- & -- & \checkmark \\
    Integration in existing SLAM pipelines & -- & -- & \checkmark \\
    \hline
    \end{tabular}
    \label{tab:related}
\end{table}

\textbf{Learning-Based Cross-Modal Alignment} --
Recently, neural approaches~\cite{li_deepi2p_2021,cattaneo_cmrnet_2019,cattaneo_cmrnext_2025} use deep architectures (e.g. CNNs) to align LiDAR depth maps with RGB images, but they typically require an approximate camera pose, are computationally heavy and generalize poorly.

Other methods estimate a pose by establishing 2D-3D correspondences and solving a PnP problem~\cite{feng_2d3d-matchnet_2019,li_2d3d-matr_2023,Mu_2025_ICCV}. However, these approaches suffer from a large domain gap: image features rely on texture and color, whereas 3D features depend on local geometry. By converting the problem into a 2D–2D domain via lightweight neural rendering, our pipeline completely avoids heavy cross-modal networks, permitting real-time execution via efficient feature detectors and matchers.

%% file: sec/3_method.tex
\section{Methodology}
\label{sec:method}

\begin{figure}[t]
    \centering
    \includegraphics[width=\linewidth]{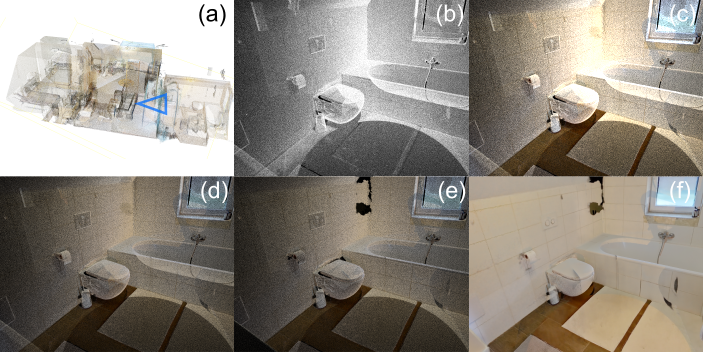}
    \caption{The steps within the adapted neural renderer. (a) First, the input point cloud and camera parameters are used to (b) compute the per-pixel minimum depth via z-buffering. In the second pass, (c) colors of points projecting to the same pixel within a small depth threshold are accumulated and then (d) averaged. Next, (e) background leakage is removed using a hierarchical depth filter, and finally, (f) any remaining holes are filled with a U-Net.}
    \label{fig:overviewrenderer}
\end{figure}

We propose two methods for camera localization with respect to a point cloud: (1) Render\&Match (R\&M) and (2) Prebuild\&Localize (P\&L).
The first method uses real-time neural rendering to synthesize a virtual image from the last known pose, which is then matched with the next camera frame. 
Our second method, P\&L, precomputes a map offline, enabling efficient, real-time tracking using the localization mode of standard SLAM pipelines.
Both approaches rely on a neural renderer for high-fidelity LiDAR point clouds~\cite{vanherck_real-time_2025}.
We choose this neural renderer because it requires no scene-specific training or dense posed images while running in real-time on commodity hardware.
The renderer projects 3D points into a 2D image plane using a hierarchical depth filter to remove background points leaking into the foreground, and employs a U-Net~\cite{ronneberger_u-net_2015} to fill gaps in the image. 
An overview of the neural rendering pipeline is shown in \Cref{fig:overviewrenderer}.

\begin{figure*}[t]
    \centering
    \includegraphics[width=\linewidth]{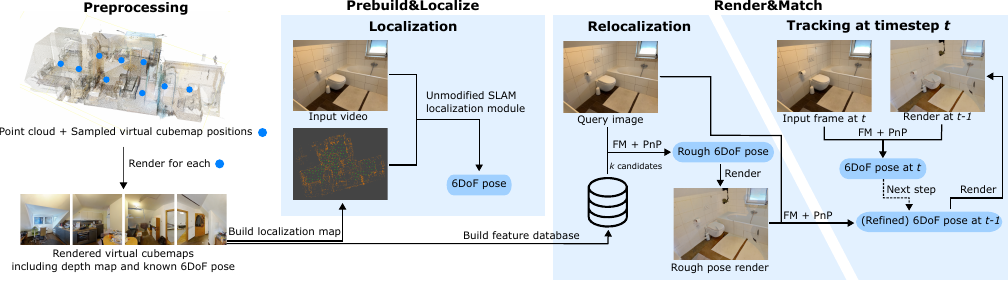}
     \caption{Overview of the proposed methods. 
    \textbf{Preprocessing:} synthetic cubemap images and depth maps are rendered from uniformly sampled poses within the LiDAR point cloud, using the neural rendering method described in \Cref{sec:neural}. \textbf{Prebuild\&Localize} utilizes these virtual cubemaps to compute a localization map offline, which can be used by the localization module of traditional SLAM backends. \textbf{Render\&Match} creates a feature database that links 2D descriptors to ground-truth 3D landmarks. \textbf{Relocalization:} when no prior pose is available, $k$ candidate keyframes selected with the global descriptor of the query image, are matched to the query to obtain a coarse 6DoF pose via PnP, which is refined by re-rendering the point cloud at the estimated pose. \textbf{Tracking:} during tracking, the previous pose is used to render a synthetic view, which is matched to the live camera frame to estimate the current 6DoF pose. Both stages operate without unbounded drift since all correspondences are derived from the LiDAR-based geometry.}
    \label{fig:overviewrm}
\end{figure*}

\subsection{Neural Point Cloud Renderer}
\label{sec:neural}

Let the LiDAR point cloud be $P = {(X_i, C_i)}$ where $X_i \in \mathbb{R}^{3}$ are 3D points and $C_i \in \mathbb{R}^{3}$ their RGB colors. Given a camera pose $T_c = [R|t] \in SE(3)$ and intrinsic matrix $K_c$, each point is projected to the image plane. Rather than relying on frustum culling and thread locking to handle cases where multiple points project to the same pixel, we divide the point cloud projection into three passes, as outlined by Sch\"utz~\etal~\cite{SCHUETZ-2021-PCC}: \textbf{(i) Depth pass --} compute per-pixel minimum depth $z_{min}(p)$ using z-buffering, \textbf{(ii) Color accumulation --} retain points within a threshold $\delta$ of $z_{min}(p)$, \textbf{(iii) Averaging --} compute per-pixel color as the average of the set of points that project to the pixel and are close to the nearest depth (within threshold $\delta$).
Background leakage is then removed by a hierarchical depth filter~\cite{vanherck_real-time_2025}, followed by a U-Net hole-filling network.
We adopt the same U-Net configuration as Vanherck \etal~\cite{vanherck_real-time_2025}, compiled using NVIDIA TensorRT for improved efficiency. 

The neural point cloud renderer outputs a hole-filled RGB image and a background-filtered depth map from a given point cloud, using the camera’s extrinsic and intrinsic parameters:
\begin{equation}
    (I_c^{\text{synth}}, D_c^{\text{synth}}) = \mathcal{R}(P, T_c, K_c).
\end{equation}

Compared to Vanherck~\etal~\cite{vanherck_real-time_2025}, our implementation eliminates the frustum-culling overhead and synchronization locks, resulting in an average speedup of $\sim$1.75$\times$ with identical visual quality for a video sequence at 960$\times$720 resolution and with 58M points. For the same resolution and 288M points, this implementation speeds up rendering by 32$\times$.

\subsection{Render\&Match (R\&M)}
\label{sec:online}

The Render\&Match method performs absolute camera localization by matching the camera feed to synthetic views of the point cloud, generated by the neural rendering method outlined in \Cref{sec:neural}.
Since each synthetic view has a known ground-truth pose and depth from the LiDAR scan, we can establish 2D-3D correspondences directly without unbounded temporal drift. 
An overview is given in \Cref{fig:overviewrm}. 
For feature extraction and matching in this pipeline, we choose XFeat~\cite{Potje_2024_CVPR} because the deep neural features are more robust to viewpoint and lighting changes compared to ORB~\cite{rublee_orb_2011}, while achieving real-time performance with GPU inference.

\textbf{Relocalization} -- To enable camera relocalization within the point cloud, we precompute a database of synthetic keyframes and global image descriptors~\cite{arandjelovic_netvlad_nodate, berton_eigenplaces_2023, berton_megaloc_nodate}. 
We designed a lightweight GUI tool that allows the user to define a region of interest within the point cloud, from which 3D positions are uniformly sampled to serve as virtual keyframe locations. The sampling interval is determined by the user, and allows for adaptibility to specific application needs, e.g. a higher sampling density in areas where more user traffic is expected. 
This is the only manual step required in the preprocessing stage. 
For each uniformly sampled 3D position, we render a cubemap with a depth map $(I_j, D_j) = \mathcal{R}(P, T_j, K_{\text{cubemap}})$ using the neural renderer, where $K_{\text{cubemap}}$ is the camera matrix used to generate a cubemap image and $T_j \in SE(3)$ the pose within the point cloud. We use cubemaps to maximize coverage in each direction, we omit up and down views as they do not contribute much to localization in typical sequences. 
From each view we extract a set of features using XFeat: $\mathcal{F}_j = \left\{(d_{jk}, X_{jk}) \mid k \in \{1,\ldots,N_j\},\ D_j(u_{jk},v_{jk})>0\right\}$, where $d_{jk}$ denotes the XFeat descriptor and $X_{jk}$ the corresponding 3D point obtained by back-projecting pixel coordinates $(u_{jk}, v_{jk})$ using depth map $D_j$:
\begin{equation}
    \label{eq:backproject}
            X_{jk} = T_{j} \cdot D_{j}^{\text{synth}}(u_{jk}, v_{jk}) \cdot K_{\text{cubemap}}^{-1}. 
    \begin{bmatrix} u_{jk} \\ v_{jk} \\ 1 \end{bmatrix}.
\end{equation}
We also compute a global image descriptor for each view, which is used to select candidate keyframes for relocalization, speeding up the relocalization process significantly.
These sets of descriptors and corresponding 3D points are stored in the feature database $\mathcal{D}$.
At runtime, $k$ candidate keyframes are selected from $\mathcal{D}$ using the global image descriptor, then features $\{d_i^q\}$ extracted from the query image $I_q$ are matched against the descriptors of the $k$ candidates.
When sufficient matches $\{m^q_i\} = \{(u^q_i, v^q_i)\}$ are found for a keyframe, the resulting correspondences $\{(m_i^q, X_i)\}$ are used to estimate a rough camera pose $T_{qr} = \text{PnP}_{\text{RANSAC}}(\{m_i^q\}, \{X_i\}, K_c)$~\cite{lepetit_epnp_2009,fischler_random_1981}.
This coarse pose is then refined by re-rendering the point cloud for $T_{qr}$ and re-estimating feature correspondences, yielding the refined pose $T_q$, which serves as the initialization for the tracking stage.
The choice of XFeat is crucial for relocalization. 
Its robustness to viewpoint changes and varying intrinsic parameters (cubemap vs. camera image) reduces the number of required cubemaps compared to methods such as ORB~\cite{rublee_orb_2011}, and enables tracking to be camera-agnostic. 
We choose EigenPlaces~\cite{berton_eigenplaces_2023} for global image descriptors due to its real-time performance, small feature dimensionality, and compact model size, while maintaining competitive accuracy. 
\Cref{tab:rm_xfeat_vs_orb} shows the impact of XFeat over ORB and the impact of the global descriptors.
We found $k=30$ to give a good balance between speed and robustness.

\textbf{Tracking} -- Given a camera image $I_t$ at timestep $t$, the camera matrix $K_c$ and the last computed 6DOF pose $T_{t-1}$:
\begin{enumerate}
    \item Render the point cloud $P$ from $T_{t-1}$ to obtain a synthetic view and depth:
    $(I_{t-1}^{\text{synth}}, D_{t-1}^{\text{synth}}) = \mathcal{R}(P, T_{t-1}, K_c).$
    \item Extract features using XFeat~\cite{Potje_2024_CVPR} in both $I_t$ and $I_{t-1}^{\text{synth}}$.
    \item Match 2D features between the camera image and the synthetic view. 
    Denote each match as a pair of pixel coordinates
    $m_k = \big((u_k^t, v_k^t),\; (u_k^{s}, v_k^{s})\big),$
    where $(u_k^t, v_k^t)$ is the pixel in the camera image $I_t$ and $(u_k^{s}, v_k^{s})$ is the corresponding pixel in the synthetic image $I_{t-1}^{\text{synth}}$.
    \item For each match, back-project the synthetic pixel $(u_k^{s}, v_k^{s})$ using the synthetic depth to obtain the 3D landmark $X_k$ (in world coordinates), similar to \cref{eq:backproject}.
    \item Estimate the 6DOF pose $T_t$ with PnP-RANSAC~\cite{fischler_random_1981, lepetit_epnp_2009} using the matches in the camera image $\{(u_k^t, v_k^t)\}$ and their corresponding 3D landmarks $\{X_k\}$.
    \item Next timestep: $T_{t-1} \leftarrow T_{t}$.
\end{enumerate}
Since each measurement stems from an accurately pre-captured LiDAR point cloud there is no unbounded accumulation of drift. The next section introduces a method that requires neural rendering only during preprocessing, making it more efficient and able to run without a powerful GPU.

\subsection{Prebuild\&Localize (P\&L)}
\label{sec:offline}

Instead of Simultaneous Localization And Mapping (SLAM), we split the process in two separate stages: mapping and localization.
For mapping, we propose to create a map directly from the LiDAR point cloud. 
Our proposed method is displayed in \Cref{fig:overviewrm}. 
We take a similar approach as in \Cref{sec:online} to achieve this.
Again, we sample 3D positions on a uniform grid within the point cloud. 
We render cubemap-like images/keyframes using the intrinsic properties of the camera $(I_j, D_j) = \mathcal{R}(P, T_j, K_{\text{cam}})$.
Because of these camera parameters, each sampled 3D position lacks full coverage. 
To improve vertical coverage, we render two views per direction, one slightly pitched upward and one downward. 
Again, straight up and down views are omitted.
From each rendered keyframe, we extract ORB features and associate them with their corresponding 3D landmarks: $\mathcal{F}_j = \left\{(d_{jk}, X_{jk}) \mid k \in \{1,\ldots,N_j\},\ D_j(u_{jk},v_{jk})>0\right\}$, where $d_{jk}$ denotes the ORB descriptor and $X_{jk}$ the reconstructed 3D landmark, analogous to \Cref{eq:backproject}. 
For SLAM map creation, landmarks observed in multiple keyframes should correspond to a single map point. For each keyframe, all landmarks are projected to the image plane. 
Feature correspondences are sought within a small 3$\times$3 pixel window around each keypoint, making matching efficient while filtering noisy LiDAR measurements. 
Matching landmarks are merged, and the map is updated. Each landmark retains the representative descriptor with the lowest median distance to its observations, as in their respective SLAM framework.
Landmarks observed by only a single keyframe are removed from the map, as they contribute little to tracking while unnecessarily increasing map density.
The resulting localization map is drift-bounded, as each keyframe and landmark is positioned using ground-truth poses obtained directly from the laser scan. 
While our approach can be applied to other visual SLAM frameworks besides ORB-SLAM3 and StellaVSLAM, these were not implemented in this work. 
In \Cref{sec:evaluation}, we show this approach improves localization accuracy compared to traditional SLAM. 

For localization, the existing SLAM backend loads the prebuilt map in localization-only mode, without requiring any modifications to the framework. 
This design eliminates the need for online rendering, resulting in significantly lower computational load (without a GPU) while maintaining real-time, drift-bounded localization.
The only preprocessing cost lies in the one-time offline generation of the localization map. 
The existing SLAM pipeline can also be leveraged to keep the map up to date with changes in the environment.

%% file: sec/4_eval.tex
\section{Evaluation}
\label{sec:evaluation}

In this section, we evaluate the performance of our localization methods. 
We conduct experiments on ScanNet++~\cite{yeshwanthliu2023scannetpp} and the Oxford Spires dataset~\cite{tao_oxford_2025}, and two custom datasets. 

\subsection{Datasets}
\textbf{ScanNet++} -- The ScanNet++~\cite{yeshwanthliu2023scannetpp} dataset provides dense LiDAR point clouds with registered camera images. 
Camera poses in ScanNet++ were obtained by running an offline COLMAP~\cite{schoenberger2016sfm} pipeline and supplementing it with rendered pseudo-images from the point cloud, through which scale can be recovered. 
These poses are refined by minimizing a dense photometric error guided by the LiDAR geometry.
Because ScanNet++ poses are computed rather than directly measured, they contain an unknown error and thus serve as an approximate reference, not a ground truth. 
Due to the extensive processing, ScanNet++ chooses to only process every 10th frame. 

\textbf{Custom datasets} -- We created two ground-truth datasets: \textbf{(i)} \textbf{Lab dataset} and \textbf{(ii) Synthetic dataset}. 
The Lab dataset is a real-world dataset using a sub-millimeter accurate motion capture system (Qualisys) and a Leica RTC360 scanner, capturing an area of $\sim$6$\times$6 m and 71 million points. Videos were captured using a Realsense D455 camera. 
We calibrate the camera to the motion-capture frame via PnP on tracked reflective targets, and align motion-capture and LiDAR frames with per-trajectory SE(3) alignment~\cite{grupp2017evo}. 
The Synthetic dataset simulates a static LiDAR scanner in Unity, covering $\sim$60$\times$40 m and 262 million points.
The videos in both datasets were captured at a resolution of 1280$\times$720px and a frame rate of 30 fps. Additional information is given in \cref{tab:custom_dataset_stats}.

\input{tables/dataset_characterization}

\textbf{Oxford Spires dataset} -- The Oxford Spires~\cite{tao_oxford_2025} dataset is a large-scale indoor and outdoor dataset, with several video sequences per environment. 
The ground-truth trajectories are computed using ICP registering the LiDAR cloud of an on-board sensor to the global LiDAR scan, acquired with a static LiDAR scanner. 
The accuracy of the ground truth trajectory is approximately 1-2 cm. 

\subsection{Experimental setting \& Evaluation Metrics}
All experiments were performed on a laptop with an Intel i7-13850HX CPU and an RTX 3500 Ada GPU. 
We implemented our Prebuild\&Localize approach for both StellaVSLAM (based on OpenVSLAM~\cite{openvslam2019}) and ORB-SLAM3~\cite{campos_orb-slam3_2021} to show our method generalizes across frameworks. 

We evaluate localization performance using the following metrics:
\textbf{(i) Pose accuracy} -- median positional error and median angular error between the estimated and ground-truth camera poses,
\textbf{(ii) Tracking robustness} -- The percentage of successfully localized frames relative to the processed frames (\%Loc.) and the percentage of tracked frames meeting accuracy thresholds (e.g. <1cm<1\textdegree). We report the average of these metrics across all sequences.

\subsection{Baselines}\label{sec:baselines}
We compare against the baselines listed in \Cref{tab:baselines}. 
Of these baselines, our methods are the only online visual localization approaches that require only a prior LiDAR map, recover metric trajectories, and do not require an initial pose. 
Our Prebuild\&Localize approach further improves accessibility by operating independently of the LiDAR point cloud, eliminating any GPU dependency during tracking.

For StellaVSLAM~\cite{openvslam2019} and ORB-SLAM3~\cite{campos_orb-slam3_2021}, we first run SLAM once to build a map with loop closing enabled. 
We then perform a second pass in pure localization mode using the same sequence, where the camera is tracked within this fixed, trajectory-optimized, pre-built map. 
This ensures a fair comparison, as all methods localize in an existing map and are not affected by loop-closing induced pose jumps, representing a best-case scenario for SLAM pipelines. 
We ran CMRNext~\cite{cattaneo_cmrnext_2025} with full trajectory localization, given an initial pose and using the last estimated pose as the coarse input for the next frame. 
For our approaches, R\&M and P\&L we create a single map for all trajectories in that scene, demonstrating that our method is not tied to a known initial pose or coarse trajectory. 
The COLMAP~\cite{schoenberger2016sfm} and VILENS~\cite{wisth_vilens_2023} trajectories are precomputed in the Oxford Spires dataset.

\input{tables/baselines}

We apply a RANSAC-based Sim(3) Umeyama alignment~\cite{grupp2017evo,fischler_random_1981} to match SLAM trajectories with the ground-truth trajectory, and RANSAC-based SE(3) Umeyama alignment for COLMAP and VILENS trajectories. 
Since ScanNet++, CMRNext and our methods already operate at the correct scale within the global LiDAR coordinate system, a main advantage is that there is no additional alignment needed for these approaches. However, for a full comparison, we also report results after aligning each method using the RANSAC-based Sim(3) alignment, denoted by `*'. These results are not considered when highlighting the best values.

\subsection{Results}
\input{tables/scannet_metrics}

\textbf{ScanNet++ --}
For each method, \Cref{tab:scannet_metrics} reports: the median positional error and median angular error with respect to the reference poses provided by the dataset, the percentage of successfully localized frames (for ScanNet++, only every 10$^{th}$ frame is processed), the percentage of frames whose pose error is below a threshold, and the Structural Similarity Index Measure (SSIM)~\cite{wang_image_2004} between the query RGB image and the corresponding neural render.
\begin{figure}[t!]
    \centering
    \includegraphics[width=\linewidth]{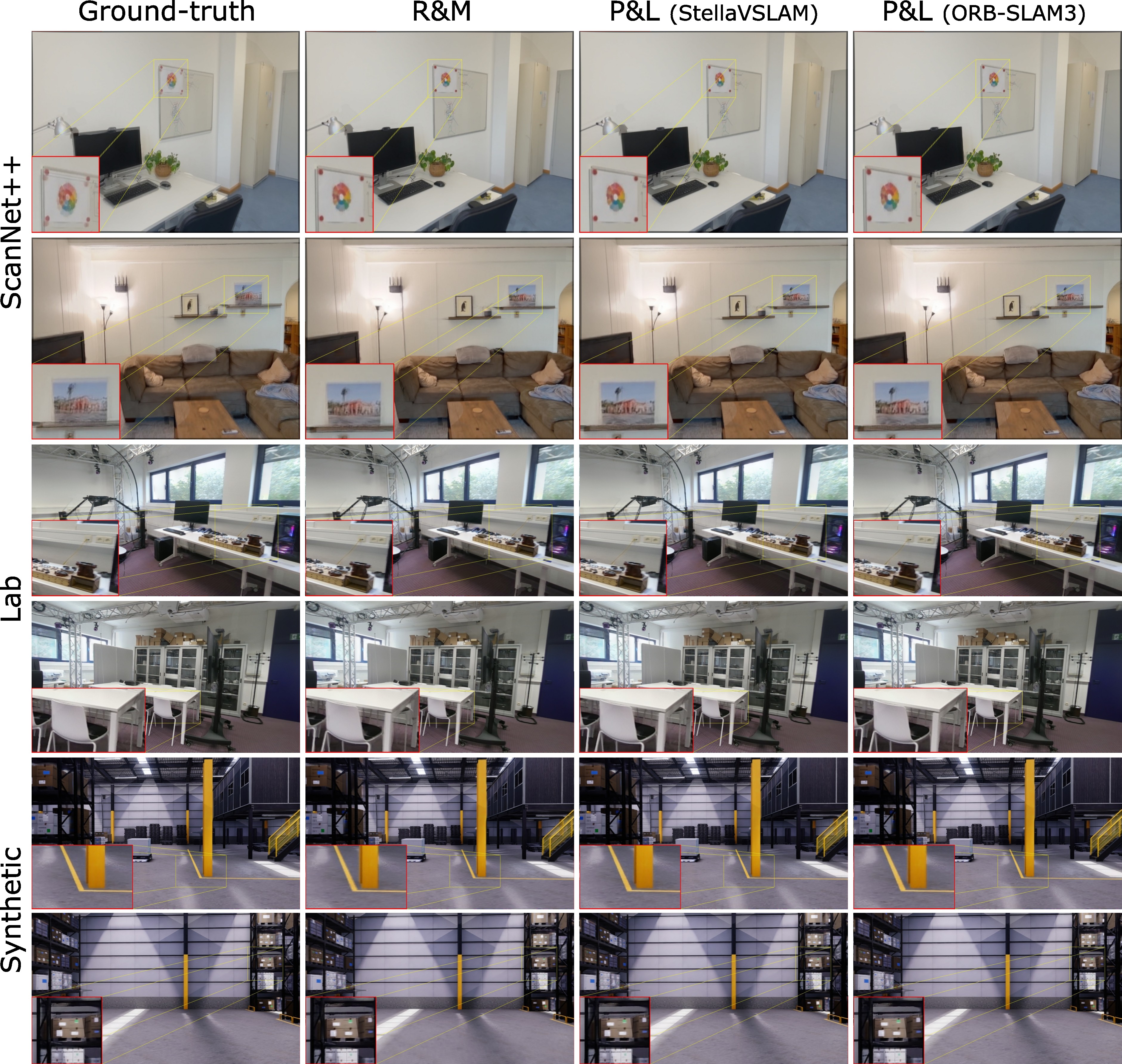}
    \caption{Results for ScanNet++ and our custom datasets. Each image displays an overlay of the target camera frame and a neural rendering of the point cloud generated from the estimated camera pose. We show results for the ground-truth, R\&M, and the P\&L methods. Misalignment, visible as ghosting or blur, suggests pose estimation errors. Note that ScanNet++'s ground-truth is only an approximate reference. }
    \label{fig:empiricalscannet}
\end{figure}
The positional and angular errors reflect local pose consistency but not absolute global accuracy, since the ScanNet++ reference poses are only approximate. 
To further evaluate camera-to-point-cloud alignment quality, we report SSIM~\cite{wang_image_2004}, which captures structural agreement between images and is less sensitive to color differences. While SSIM directly measures photometric and render-level consistency rather than direct 3D geometric accuracy, higher structural agreement serves as a proxy for successful image alignment. 
Our approaches significantly outperform the standard SLAM frameworks in localization accuracy. 
Furthermore, as demonstrated by higher SSIM scores across a larger number of successfully estimated frames, our method achieves superior render-to-image alignment compared to the ScanNet++ poses.
We further show this in \Cref{fig:empiricalscannet}, which illustrates examples of rendered views overlaid with real camera images, where visible ghosting or blur indicates pose misalignment. 
The P\&L variant of ORB-SLAM3 tracks significantly fewer frames than that of StellaVSLAM because its relocalization strategy is less robust, as also reported in \cref{tab:relocalization}. 
In particular, StellaVSLAM matches neighboring keyframes of relocalization candidates, whereas ORB-SLAM3 does not.

\input{tables/custom_metrics}

\input{tables/decimation_results}

\textbf{Custom dataset evaluation --} 
\Cref{tab:custom_metrics} compares StellaVSLAM, ORB-SLAM3, our enhanced variants (P\&L) and the Render\&Match approach in terms of median positional error (cm) and median angular error ($^{\circ}$), the percentage of successfully localized frames, and the percentage of tracked frames under specified accuracy thresholds. 
The results show that our method increases accuracy, while maintaining similar robustness.

\Cref{tab:decimation_results} reports the localization accuracy on the `lab' dataset under point cloud decimation for the R\&M and P\&L approaches, using neural and traditional point-based rendering, as well as the intermediate depth-filtered renders. 
For Prebuild\&Localize, we only evaluate StellaVSLAM due to its superior relocalization strategy.

\begin{figure}[b!]
    \centering
    \includegraphics[width=\linewidth]{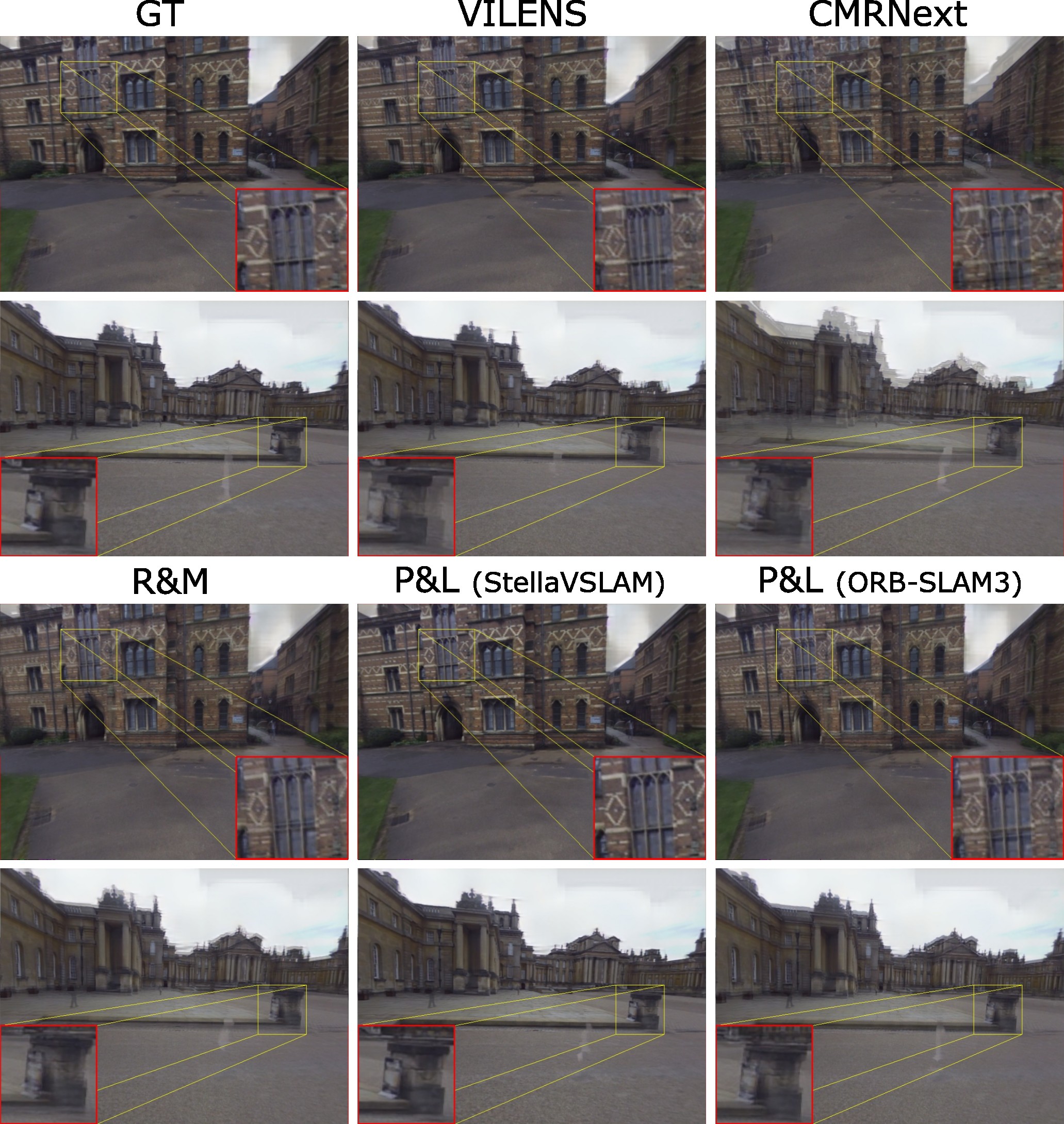}
    \caption{Results on the Oxford Spires dataset. Each image displays an overlay of the target camera frame and a neural rendering of the point cloud generated from the estimated camera pose. We show results for the ground-truth, VILENS~\cite{wisth_vilens_2023}, CMRNext~\cite{cattaneo_cmrnext_2025}, R\&M, and the P\&L methods. Misalignment, visible as ghosting or blur, suggests pose estimation errors.}
    \label{fig:empiricaloxford}
\end{figure}

The original point cloud contains 71M points and is sufficiently dense for point-based and depth-filtered rendering to achieve strong localization without decimation.
However, as the point cloud is further decimated, accuracy degrades and eventually tracking fails.
In contrast, our neural rendering approach undergoes a graceful degradation and keeps very strong localization under the 5cm,2$^\circ$ threshold.
\\

\textbf{Oxford Spires evaluation --}
Results are reported in \Cref{tab:oxford_metrics} and \Cref{fig:empiricaloxford}.
We show the median positional and median angular error with respect to the ground truth trajectory, the percentage of tracked frames, and the percentage of tracked frames below defined accuracy thresholds. 
Our method achieves accuracy comparable to the offline COLMAP pipeline while enabling real-time camera registration.

CMRNext, trained mainly on outdoor autonomous-driving data, does not require a dense colored point cloud, but performs poorly indoors, with only 31.5\% of tracked frames localized within 20 cm. 
StellaVSLAM tracks the most frames but achieves the lowest accuracy. 
VILENS achieves high accuracy, yet requires additional hardware: camera, IMU and LiDAR. 
In contrast, our method performs online, purely monocular localization within a prior LiDAR map.

The P\&L variants track fewer frames than their SLAM counterparts, as our map is not optimized for the specific evaluation trajectory, while SLAM methods jointly optimize map and motion. 
Taking into account a rough prior trajectory during map construction could further improve tracking results.

\input{tables/oxford_metrics}
\begin{figure}[b]
    \centering
    \includegraphics[width=\linewidth]{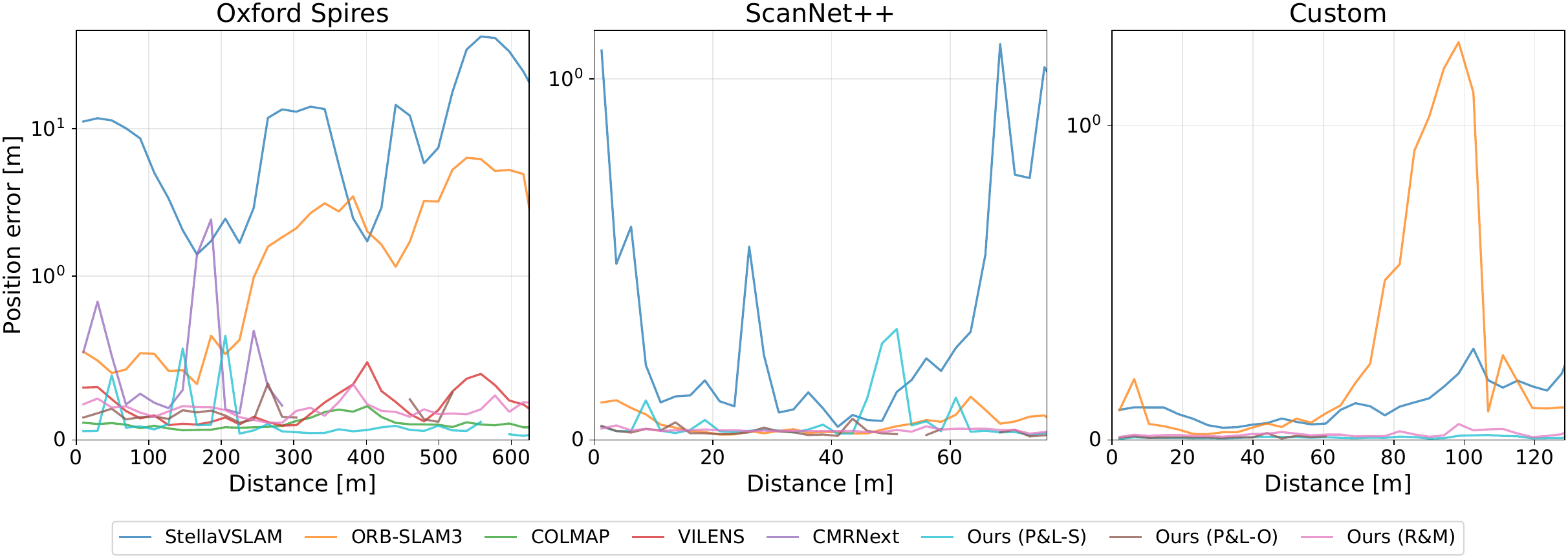}
    \caption{Error-vs-distance plot for every baseline and our methods, pooled across all sequences of each dataset and binned into 40 intervals, each curve shows the per-bin mean. Every estimated trajectory is aligned with the ground truth as described in \cref{sec:baselines}. }
    \label{fig:errorvsdistance}
\end{figure}

\textbf{Error-vs-distance --} \Cref{fig:errorvsdistance} shows an error-vs-distance plot for each dataset. This plot shows increasing errors for the baseline SLAM methods, while our methods, COLMAP, VILENS and CMRNext do not or to a lesser extent, indicating that our methods are less prone to accumulating drift. The SLAM baselines have loop-closing enabled, making traditional drift accumulation less visible. 

\textbf{Relocalization analysis --} \Cref{tab:relocalization} shows the relocalization success rate and time-to-first-pose across 50 evenly spaced cold starts per scene. Relocalization is successful if a pose within 50 cm is estimated within 10 seconds, while each attempt is allowed to run for 30 seconds. It is important to note that the baseline SLAM methods use a map that was created with the exact same video sequence, meaning the map is fully optimized for the trajectory, in contrast to our methods. While the P\&L methods employ the respective, built-in relocalization strategies of their SLAM systems, our R\&M approach achieves high performance by combining global image descriptors with robust XFeat features. As discussed, the P\&L-O variant relocalizes fewer frames because its relocalization strategy is less robust than that of P\&L-S.

\input{tables/reloc_experiment}

\textbf{Runtime analysis --} We report a runtime performance analysis in \Cref{tab:timings}.
With the exception of CMRNext, the measured runtimes strongly depend on the number of landmarks and keyframes in the constructed map, as well as on the size of the point cloud, which affects the load of the neural renderer. 
Although map construction adds computational overhead, the system achieves real-time performance during tracking.

\input{tables/timings}
\input{tables/rm_xfeat_vs_orb}

\textbf{R\&M variants --} \Cref{tab:rm_xfeat_vs_orb} shows the results of the R\&M pipeline using XFeat features (with and without global image descriptors) and ORB features. 
The ORB variant tracks significantly fewer frames, supporting our choice of XFeat, especially since the approach already requires GPU execution. 
While global image descriptors speed up relocalization by selecting candidate keyframes, tracking performance decreases because the optimal keyframe is not always included among the candidates. 
The variant without global image descriptors uses all available keyframes as a relocalization candidate.

%% file: tables/dataset_characterization.tex
\begin{table}[h]
    \centering
    \caption{Custom dataset characterization (mean $\pm$ std over sequences). \#Scans reports the number of concatenated scans.}
    \label{tab:custom_dataset_stats}
    \resizebox{\columnwidth}{!}
    {
    \setlength{\tabcolsep}{2pt}
    \begin{tabular}{lcccccc}
        \hline
        Scene & \#Scans & \#Seq. & \#Frames & Length & Trans. speed & Ang. speed \\
         &  &  &  & (m) & (m/s) & ($^\circ$/s) \\
        \hline
        Lab       & 4 & 3 & $4335\pm102.3$ & $45.6\pm12.9$ & $0.32\pm0.08$ & $23.7\pm7.5$ \\
        Synthetic & 29 & 4 & $1968\pm265.2$ & $134.2\pm24.1$ & $2.04\pm0.12$ & $10.4\pm1.4$ \\
        \hline
    \end{tabular}
    }
\end{table}

%% file: tables/baselines.tex
\begin{table}[htb]
\caption{Comparison of baseline methods. Columns indicate operation mode (Type: Online [On] / Offline [Off]), required sensor modalities, GPU independence (-GPU), metric trajectory recovery (Met.), and relocalization support (Rel.).}
\label{tab:baselines}
\centering
\resizebox{\columnwidth}{!}{%
\begin{tabular}{l|c|l|c|c|c}
\hline
\textbf{Method} & \textbf{Type} & \textbf{Data Required} & \textbf{-GPU} & \textbf{Met.} & \textbf{Rel.} \\ 
\hline
ScanNet++~\cite{yeshwanthliu2023scannetpp} & Off & Vis.+RGB LiDAR & \checkmark & \checkmark & NA \\
StellaVSLAM~\cite{openvslam2019}            & On  & Vis. (Monocular)      & \checkmark & --         & \checkmark \\
ORB-SLAM3~\cite{campos_orb-slam3_2021}      & On  & Vis. (Monocular)      & \checkmark & --         & \checkmark \\
COLMAP~\cite{schoenberger2016sfm}           & Off & Vis. (Monocular)      & \checkmark & --         & NA \\
VILENS~\cite{wisth_vilens_2023}             & On  & Vis.+IMU+LiDAR        & \checkmark & \checkmark & \checkmark \\
CMRNext~\cite{cattaneo_cmrnext_2025}        & On  & Vis.+LiDAR            & --         & \checkmark & -- \\
\hline
Ours (R\&M)                                & On  & Vis.+RGB LiDAR        & --         & \checkmark & \checkmark \\
Ours (P\&L)                                 & On  & Vis.+RGB LiDAR        & \checkmark & \checkmark & \checkmark \\
\hline
\end{tabular}%
}
\end{table}

%% file: tables/scannet_metrics.tex
\begin{table}[b]
\caption{Quantitative evaluation on ScanNet++. The \highlight{best}{best} and \highlight{secondbest}{second-best} values are highlighted.}\label{tab:scannet_metrics}
\centering
\resizebox{\columnwidth}{!}{
\small
\begin{tabular}{l|cc|c|c|c}
\hline
\textbf{Method} & \textbf{Pos.} & \textbf{Ang.} & \textbf{\%Loc.} & \textbf{<5cm} & \textbf{SSIM} \\
 & \textbf{(m)$\downarrow$} & \textbf{($^{\circ}$)$\downarrow$} & \textbf{$\uparrow$} & \textbf{<2$^{\circ}\uparrow$} & \textbf{$\uparrow$} \\
\hline
ScanNet++~\cite{yeshwanthliu2023scannetpp} & - & - & 86.1 & - & 0.663 \\
StellaVSLAM~\cite{openvslam2019} & 0.13 & 1.79 & 66.3 & 49.74 & 0.607 \\
ORB-SLAM3~\cite{campos_orb-slam3_2021} & 0.03 & 6.99 & \cellcolor{secondbest}70.9 & 64.04 & 0.609 \\
\hline
Ours (P\&L-S) & \cellcolor{best}0.02 & \cellcolor{best}0.44 & 69.1 & \cellcolor{secondbest}89.23 & \cellcolor{secondbest}0.691 \\
Ours (P\&L-O) & \cellcolor{secondbest}0.02 & \cellcolor{secondbest}0.50 & 31.3 & \cellcolor{best}89.50 & \cellcolor{best}0.715 \\
Ours (R\&M) & 0.02 & 0.57 & \cellcolor{best}81.6 & 87.73 & 0.682 \\
\hline
Ours (P\&L-S)* & 0.02 & 0.47 & 69.1 & 89.72 & 0.666 \\
Ours (P\&L-O)* & 0.02 & 0.58 & 31.3 & 89.85 & 0.685 \\
Ours (R\&M)* & 0.02 & 0.58 & 81.6 & 87.55 & 0.668 \\
\hline
\end{tabular}
}
\end{table}

%% file: tables/custom_metrics.tex
\begin{table}[b]
\caption{Quantitative evaluation on the custom datasets. The \highlight{best}{best} and \highlight{secondbest}{second-best} values are highlighted.}
\label{tab:custom_metrics}
\centering
\resizebox{\columnwidth}{!}{
\begin{tabular}{l|cc|c|ccc}
\hline
\textbf{Method} & \textbf{Pos.} & \textbf{Ang.} & \textbf{\%Loc.} & \textbf{<1cm} & \textbf{<5cm} & \textbf{<20cm} \\
 & \textbf{(cm)$\downarrow$} & \textbf{($^{\circ}$)$\downarrow$} & \textbf{$\uparrow$} & \textbf{<1$^{\circ}\uparrow$} & \textbf{<2$^{\circ}\uparrow$} & \textbf{<3$^{\circ}\uparrow$} \\
\hline
StellaVSLAM~\cite{openvslam2019} & 7.11 & 0.82 & 84.8 & 31.9 & 44.4 & 82.6 \\
ORB-SLAM3~\cite{campos_orb-slam3_2021} & 29.70 & 5.00 & 82.3 & 34.2 & 50.4 & 75.2 \\
\hline
Ours (P\&L-S) & \cellcolor{best}0.59 & \cellcolor{best}0.18 & \cellcolor{best}100.0 & \cellcolor{best}75.4 & \cellcolor{best}98.6 & \cellcolor{best}99.9 \\
Ours (P\&L-O) & 1.53 & 0.50 & 39.2 & \cellcolor{secondbest}49.3 & 82.2 & 92.5 \\
Ours (R\&M) & \cellcolor{secondbest}1.21 & \cellcolor{secondbest}0.21 & \cellcolor{secondbest}97.4 & 45.0 & \cellcolor{secondbest}92.4 & \cellcolor{secondbest}99.7 \\
\hline
Ours (P\&L-S)* & 0.56 & 0.18 & 100.0 & 77.3 & 98.7 & 99.9 \\
Ours (P\&L-O)* & 0.94 & 0.62 & 39.2 & 56.1 & 84.8 & 99.5 \\
Ours (R\&M)* & 1.22 & 0.21 & 97.4 & 44.2 & 92.5 & 99.7 \\
\hline
\end{tabular}}
\end{table}

%% file: tables/decimation_results.tex
\begin{table}[htb]
      \caption{Impact of scene decimation on localization accuracy for the `lab' dataset, comparing Neural, Depth-filtered and Point-based rendering. Results are shown by method, with the superior performance \highlight{secondbest}{highlighted} per decimation level.}
\label{tab:decimation_results}
\centering
\setlength{\tabcolsep}{2pt}
\resizebox{\columnwidth}{!}{
\begin{tabular}{cc|ccc|ccc|ccc}
\hline
& & \multicolumn{3}{c|}{\textbf{Neural}} & \multicolumn{3}{c|}{\textbf{Depth-filtered}} & \multicolumn{3}{c}{\textbf{Point-based}} \\
\cline{3-11}
\textbf{Method} & \textbf{Dec.} & \textbf{Pos.$\downarrow$} & \textbf{\%Loc.$\uparrow$} & \textbf{<5cm,<2$^{\circ}\uparrow$} & \textbf{Pos.$\downarrow$} & \textbf{\%Loc.$\uparrow$} & \textbf{<5cm,<2$^{\circ}\uparrow$} & \textbf{Pos.$\downarrow$} & \textbf{\%Loc.$\uparrow$} & \textbf{<5cm,<2$^{\circ}\uparrow$} \\
\hline
 R\&M & 100\% & \cellcolor{secondbest}0.77 & \cellcolor{secondbest}94.0 & \cellcolor{secondbest}98.6 & 1.90 & 86.1 & 88.0 & 1.02 & 82.6 & 96.9 \\
 & 50\% & \cellcolor{secondbest}0.88 & \cellcolor{secondbest}88.9 & \cellcolor{secondbest}98.6 & 3.15 & 73.0 & 71.7 & 1.89 & 75.2 & 87.8 \\
 & 20\% & \cellcolor{secondbest}0.98 & \cellcolor{secondbest}83.6 & \cellcolor{secondbest}98.3 & 4.98 & 35.9 & 50.6 & 3.67 & 29.6 & 63.3 \\
 & 10\% & \cellcolor{secondbest}1.08 & \cellcolor{secondbest}83.7 & \cellcolor{secondbest}97.9 & 5.72 & 7.1 & 32.7 & 11.20 & 2.4 & 30.6 \\
 & 5\% & \cellcolor{secondbest}1.27 & \cellcolor{secondbest}83.5 & \cellcolor{secondbest}97.0 & - & - & - & - & - & - \\
 & 2\% & \cellcolor{secondbest}1.69 & \cellcolor{secondbest}82.4 & \cellcolor{secondbest}93.5 & - & - & - & - & - & - \\
\hline
P\&L & 100\% & \cellcolor{secondbest}0.40 & \cellcolor{secondbest}100.0 & \cellcolor{secondbest}99.6 & 0.42 & \cellcolor{secondbest}100.0 & 99.4 & 1.59 & 91.2 & 76.5 \\
 & 50\% & \cellcolor{secondbest}0.42 & \cellcolor{secondbest}100.0 & \cellcolor{secondbest}99.5 & 0.50 & 98.6 & 99.2 & 152.52 & 83.7 & 0.0 \\
 & 20\% & \cellcolor{secondbest}0.50 & \cellcolor{secondbest}98.4 & \cellcolor{secondbest}99.4 & 0.91 & 77.5 & 96.5 & 8.63 & 42.5 & 46.1 \\
 & 10\% & \cellcolor{secondbest}0.79 & \cellcolor{secondbest}80.4 & \cellcolor{secondbest}98.8 & 1.42 & 10.2 & 48.1 & 34.03 & 18.0 & 0.0 \\
 & 5\% & \cellcolor{secondbest}0.91 & \cellcolor{secondbest}17.2 & \cellcolor{secondbest}97.4 & - & - & - & - & - & - \\
 & 2\% & - & - & - & - & - & - & - & - & - \\
\hline
\end{tabular}}

\end{table}

%% file: tables/oxford_metrics.tex
\begin{table}[t]
\caption{Quantitative evaluation on the Oxford Spires dataset. The \highlight{best}{best} and \highlight{secondbest}{second-best} values are highlighted.}
\label{tab:oxford_metrics}
\centering
\resizebox{\columnwidth}{!}{
\small
\setlength{\tabcolsep}{2.2pt}
\begin{tabular}{l|cc|c|ccc|ccc}
\hline
\textbf{Method} & \textbf{Pos.} & \textbf{Ang.} & \textbf{\%Loc.} & \multicolumn{3}{c|}{\textbf{Pos. $\uparrow$}} & \multicolumn{3}{c}{\textbf{Ang. $\uparrow$}} \\
 & \textbf{(m)$\downarrow$} & \textbf{($^{\circ}$)$\downarrow$} & \textbf{$\uparrow$} & \textbf{<5c} & \textbf{<20c} & \textbf{<1m} & \textbf{<1$^{\circ}$} & \textbf{<2$^{\circ}$} & \textbf{<3$^{\circ}$} \\
\hline
StellaVSLAM~\cite{openvslam2019} & 2.02 & 1.61 & \cellcolor{best}85.2 & 15.9 & 36.4 & 55.4 & 56.3 & 81.7 & 84.4 \\
ORB-SLAM3~\cite{campos_orb-slam3_2021} & 0.52 & 16.71 & 69.5 & 31.3 & 55.3 & 70.9 & 84.5 & 86.8 & 88.7 \\
COLMAP~\cite{schoenberger2016sfm} & \cellcolor{secondbest}0.09 & - & 5.2$^\dagger$ & 31.2 & \cellcolor{secondbest}90.1 & \cellcolor{best}100.0 & - & - & - \\
VILENS~\cite{wisth_vilens_2023} & 0.13 & \cellcolor{secondbest}0.40 & 5.3$^\dagger$ & \cellcolor{secondbest}35.0 & 73.4 & 95.1 & 87.1 & 98.1 & \cellcolor{best}99.9 \\
CMRNext~\cite{cattaneo_cmrnext_2025} & 1.62 & 14.39 & 29.4 & 1.3 & 31.5 & 83.5 & 13.6 & 32.4 & 43.5 \\
\hline
Ours (P\&L-S) & \cellcolor{best}0.05 & \cellcolor{best}0.33 & 43.3 & \cellcolor{best}51.3 & \cellcolor{best}94.7 & 98.5 & \cellcolor{best}93.3 & 97.5 & 98.5 \\
Ours (P\&L-O) & 0.12 & 0.75 & 22.8 & 3.5 & 81.5 & \cellcolor{secondbest}99.8 & \cellcolor{secondbest}90.5 & \cellcolor{best}99.5 & \cellcolor{secondbest}99.9 \\
Ours (R\&M) & 0.13 & 0.83 & \cellcolor{secondbest}76.3 & 8.7 & 69.9 & 99.6 & 79.6 & \cellcolor{secondbest}98.2 & 99.5 \\
\hline
COLMAP~\cite{schoenberger2016sfm}* & 0.05 & - & 5.2$^\dagger$ & 56.4 & 98.5 & 100.0 & - & - & - \\
VILENS~\cite{wisth_vilens_2023}* & 0.12 & 0.40 & 5.3$^\dagger$ & 40.8 & 74.3 & 94.7 & 87.1 & 98.0 & 99.9 \\
CMRNext~\cite{cattaneo_cmrnext_2025}* & 0.19 & 20.25 & 29.4 & 9.1 & 55.7 & 93.2 & 5.8 & 15.4 & 30.7 \\
Ours (P\&L-S)* & 0.04 & 0.42 & 43.3 & 70.1 & 95.6 & 98.5 & 87.6 & 95.5 & 98.1 \\
Ours (P\&L-O)* & 0.08 & 0.82 & 22.8 & 28.8 & 87.9 & 99.8 & 81.0 & 99.5 & 99.9 \\
Ours (R\&M)* & 0.10 & 0.82 & 76.3 & 27.4 & 78.1 & 99.7 & 80.6 & 98.2 & 99.5 \\
\hline
\multicolumn{10}{l}{ $^{\dagger}$Available reference poses, should not be used to compare method robustness.}
\end{tabular}}
\end{table}

%% file: tables/reloc_experiment.tex
\begin{table}[t]
\centering
\caption{Global relocalization: success rate (\%, $\leq$50\,cm within 10\,s) and median time-to-first-pose (s).}
\label{tab:relocalization}
\resizebox{\linewidth}{!}{
\begin{tabular}{lcccccc}
\hline
\textbf{Method} & \multicolumn{2}{c}{\textbf{Oxford Spires}} & \multicolumn{2}{c}{\textbf{ScanNet++}} & \multicolumn{2}{c}{\textbf{Custom}} \\
\cline{2-7}
 & Succ.$\uparrow$ & t2p$\downarrow$ & Succ.$\uparrow$ & t2p$\downarrow$ & Succ.$\uparrow$ & t2p$\downarrow$ \\
\hline
StellaVSLAM & 88 & 0.08 & 78 & 0.03 & 81 & 0.05 \\
ORB-SLAM3 & 76 & 0.08 & 65 & 0.03 & 85 & 0.06 \\
Ours (R\&M) & 69 & 0.51 & 92 & 1.17 & 93 & 0.06 \\
Ours (P\&L-S) & 23 & 8.86 & 82 & 0.81 & 80 & 0.17 \\
Ours (P\&L-O) & 7 & 11.3 & 50 & 4.56 & 36 & 0.71 \\
\hline
\end{tabular}}
\end{table}

%% file: tables/timings.tex
\begin{table}[b]
\caption{Average tracking FPS, relocalization time, and map construction time across all datasets.}
\label{tab:timings}
\centering
\small
\setlength{\tabcolsep}{3.5pt} 
\begin{tabular}{l|c|c|c}
\hline
\textbf{Method} & \textbf{FPS~$\uparrow$} & \textbf{Reloc. (ms)~$\downarrow$} & \textbf{Map (s)~$\downarrow$} \\
\hline
StellaVSLAM~\cite{openvslam2019} & 45.5 & 41 & - \\
ORB-SLAM3~\cite{campos_orb-slam3_2021} & 32.5 & 23 & - \\
CMRNext~\cite{cattaneo_cmrnext_2025} & 1.6 & - & - \\
\hline
Ours (P\&L-Stella) & 33.2 & 72 & 449 \\
Ours (P\&L-ORB-SLAM3) & 31.9 & 30 & 460 \\
Ours (R\&M) & 13.5 & 47 & 83 \\
\hline
\end{tabular}
\end{table}

%% file: tables/rm_xfeat_vs_orb.tex
\begin{table}[b]
\caption{Comparison of R\&M feature backend variants. Mean metrics are reported across all datasets.}
\label{tab:rm_xfeat_vs_orb}
\centering
\small
\setlength{\tabcolsep}{6pt}
\begin{tabular}{l|c|c|c}
\hline
\textbf{Method} & \textbf{\%Loc.~$\uparrow$} & \textbf{Reloc.~(ms)~$\downarrow$} & \textbf{Map~(s)~$\downarrow$} \\
\hline
R\&M XFeat & 83.8 & 47.2 & 83.0 \\
R\&M XFeat$^{\dagger}$ & 89.0 & 659.8 & 69.0 \\
R\&M ORB & 27.5 & 45.9 & 52.8 \\
\hline
\multicolumn{4}{l}{\scriptsize $^{\dagger}$Without global image descriptors for relocalization.}
\end{tabular}
\end{table}

%% file: sec/5_conclusion.tex
\section{Conclusion \& Future Work}
We presented a real-time camera-to-LiDAR registration framework using neural rendering to bridge the domain gap between renders and real camera images. Both proposed variants (Render\&Match and Prebuild\&Localize) outperform the baselines while tracking in a true metric coordinate system. Prebuild\&Localize generates drift-bounded maps offline, allowing standard visual SLAM pipelines to track cameras without framework modifications or active GPU rendering.

Integrating our system into standard SLAM frameworks provides broad utility, especially for robotics and embedded-device localization, enabling accurate metric-scale tracking from the first frame.

Several directions are promising for future work. 
First, varying the lighting conditions of synthetic renders could further enhance the robustness of tracking. 
Second, our current method relies on a pre-acquired LiDAR scan of the environment, which may limit tracking if the environment undergoes significant changes. 
A potential solution is dynamic map updating: fixed LiDAR-based landmarks could maintain stability, while new features are added online by a mapping module to account for environmental changes. 

%% file: bib/main.bib
@String(CVPR= {IEEE Conf. Comput. Vis. Pattern Recog.})

@String(ICCV= {Int. Conf. Comput. Vis.})

@String(ECCV= {Eur. Conf. Comput. Vis.})

@String(CVPR  = {CVPR})

@String(ICCV  = {ICCV})

@String(ECCV  = {ECCV})

@article{munoz-salinas_lidar_2025,
  title={LiDAR as a Geometric Prior: Enhancing Camera Pose Tracking Through High-Fidelity View Synthesis},
  author={Muñoz-Salinas, Rafael and Liu, Jianheng and Romero-Ramirez, Francisco J. and Marín-Jiménez, Manuel J. and Zhang, Fu},
  journal={Applied Sciences},
  volume={15},
  number={15},
  pages={8743},
  year={2025}
}

@inproceedings{cattaneo_cmrnet_2019,
  title={CMRNet: Camera to LiDAR-Map Registration},
  author={Cattaneo, D. and Vaghi, M. and Ballardini, A. L. and Fontana, S. and Sorrenti, D. G. and Burgard, W.},
  booktitle={ITSC},
  pages={1283--1289},
  year={2019}
}

@inproceedings{caselitz_monocular_2016,
  title={Monocular camera localization in 3D LiDAR maps},
  author={Caselitz, Tim and Steder, Bastian and Ruhnke, Michael and Burgard, Wolfram},
  booktitle={IROS},
  pages={1926--1931},
  year={2016}
}

@inproceedings{bai_colmap-pcd_2024,
  title={Colmap-PCD: An Open-source Tool for Fine Image-to-point cloud Registration},
  author={Bai, Chunge and Fu, Ruijie and Gao, Xiang},
  booktitle={ICRA},
  pages={1723--1729},
  year={2024}
}

@inproceedings{Potje_2024_CVPR,
  author={Potje, Guilherme and Cadar, Felipe and Araujo, Andr\'e and Martins, Renato and Nascimento, Erickson R.},
  title={XFeat: Accelerated Features for Lightweight Image Matching},
  booktitle={CVPR},
  pages={2682--2691},
  year={2024}
}

@inproceedings{li_2d3d-matr_2023,
  title={2D3D-MATR: 2D-3D Matching Transformer for Detection-free Registration between Images and Point Clouds},
  author={Li, Minhao and Qin, Zheng and Gao, Zhirui and Yi, Renjiao and Zhu, Chenyang and Guo, Yulan and Xu, Kai},
  booktitle={ICCV},
  pages={14128--14138},
  year={2023}
}

@inproceedings{khatib_gsvisloc_2025,
  title={Generalizable Visual Localization for Gaussian Splatting Scene Representations},
  author={Khatib, Fadi and Moran, Dror and Trostianetsky, Guy and Kasten, Yoni and Galun, Meirav and Basri, Ronen},
  booktitle={ICCV Workshops},
  pages={178--189},
  year={2025}
}

@inproceedings{panek_meshloc_2022,
  title={MeshLoc: Mesh-Based Visual Localization},
  author={Panek, Vojtech and Kukelova, Zuzana and Sattler, Torsten},
  booktitle={ECCV},
  pages={589--609},
  year={2022}
}

@article{zheng_fast-livo2_2025,
  title={FAST-LIVO2: Fast, Direct LiDAR-Inertial-Visual Odometry},
  author={Zheng, Chunran and Xu, Wei and Zou, Zuhao and Hua, Tong and Yuan, Chongjian and He, Dongjiao and Zhou, Bingyang and Liu, Zheng and Lin, Jiarong and Zhu, Fangcheng and Ren, Yunfan and Wang, Rong and Meng, Fanle and Zhang, Fu},
  journal={IEEE Transactions on Robotics},
  volume={41},
  pages={326--346},
  year={2025}
}

@article{singh_evaluation_2025,
  title={Evaluation of XR device's real-world tracking accuracy and depth perception from an industrial point of view},
  author={Singh, Shubham and Sharma, Yash and Liaqat, Amer and Kalawsky, Roy S.},
  journal={Virtual Reality},
  volume={29},
  number={3},
  pages={118},
  year={2025}
}

@inproceedings{feng_2d3d-matchnet_2019,
  title={2D3D-MatchNet: Learning to Match Keypoints Across 2D Image and 3D Point Cloud},
  author={Feng, Mengdan and Hu, Sixing and Ang, Marcelo H. and Lee, Gim Hee},
  booktitle={ICRA},
  pages={4790--4796},
  year={2019}
}

@inproceedings{yeshwanthliu2023scannetpp,
  title={ScanNet++: A High-Fidelity Dataset of 3D Indoor Scenes},
  author={Yeshwanth, Chandan and Liu, Yueh-Cheng and Nie{\ss}ner, Matthias and Dai, Angela},
  booktitle = {ICCV},
  year={2023},
  pages={12--22}
}

@inproceedings{kim_stereo_2018,
  title={Stereo Camera Localization in 3D LiDAR Maps},
  author={Kim, Youngji and Jeong, Jinyong and Kim, Ayoung},
  booktitle={IROS},
  pages={5826--5833},
  year={2018}
}

@inproceedings{yu_monocular_2020,
  title={Monocular Camera Localization in Prior LiDAR Maps with 2D-3D Line Correspondences},
  author={Yu, Huai and Zhen, Weikun and Yang, Wen and Zhang, Ji and Scherer, Sebastian},
  booktitle={IROS},
  pages={4588--4594},
  year={2020}
}

@inproceedings{mur-artal_fast_2014,
  title={Fast Relocalisation and Loop Closing in Keyframe-Based SLAM},
  author={Mur-Artal, Raul and Tardós, Juan D.},
  booktitle={ICRA},
  pages={846--853},
  year={2014}
}

@inproceedings{vanherck_real-time_2025,
  title={Real-time Neural Rendering of LiDAR Point Clouds},
  author={Vanherck, Joni and Zoomers, Brent and Mertens, Tom and Jorissen, Lode and Michiels, Nick},
  booktitle={Eurographics 2025 - Short Papers},
  year={2025}
}

@article{fischler_random_1981,
	title = {Random sample consensus: a paradigm for model fitting with applications to image analysis and automated cartography},
	volume = {24},
	issn = {0001-0782},
	shorttitle = {Random sample consensus},
	number = {6},
	journal = {Communications of the ACM},
	author = {Fischler, Martin A. and Bolles, Robert C.},
	year = {1981},
	pages = {381--395},
}

@article{lepetit_epnp_2009,
	title = {{EPnP}: {An} {Accurate} {O}(n) {Solution} to the {PnP} {Problem}},
	volume = {81},
	issn = {0920-5691, 1573-1405},
	shorttitle = {{EPnP}},
	language = {en},
	number = {2},
	journal = {International Journal of Computer Vision},
	author = {Lepetit, Vincent and Moreno-Noguer, Francesc and Fua, Pascal},
	year = {2009},
	pages = {155--166},
}

@inproceedings{schoenberger2016sfm,
  title={Structure-from-Motion Revisited},
  author={Sch{\"o}nberger, Johannes Lutz and Frahm, Jan-Michael},
  booktitle={CVPR},
  year={2016}
}

@article{garrido-jurado_automatic_2014,
  title={Automatic generation and detection of highly reliable fiducial markers under occlusion},
  author={Garrido-Jurado, S. and Muñoz-Salinas, R. and Madrid-Cuevas, F.J. and Marín-Jiménez, M.J.},
  journal={Pattern Recognition},
  volume={47},
  number={6},
  pages={2280--2292},
  year={2014}
}

@inproceedings{openvslam2019,
  title={OpenVSLAM: A Versatile Visual SLAM Framework},
  author={Sumikura, Shinya and Shibuya, Mikiya and Sakurada, Ken},
  booktitle={MM '19},
  pages={2292--2295},
  year={2019}
}

@inproceedings{ronneberger_u-net_2015,
  title={U-Net: Convolutional Networks for Biomedical Image Segmentation},
  author={Ronneberger, Olaf and Fischer, Philipp and Brox, Thomas},
  booktitle={MICCAI},
  year={2015}, 
  pages={234--241},
}

@article{wang_image_2004,
  title={Image Quality Assessment: From Error Visibility to Structural Similarity},
  author={Wang, Zhou and Bovik, A.C. and Sheikh, H.R. and Simoncelli, E.P.},
  journal={IEEE Transactions on Image Processing},
  volume={13},
  number={4},
  pages={600--612},
  year={2004}
}

@inproceedings{pointersect,
title = {Pointersect: Neural Rendering with Cloud-Ray Intersection},
booktitle = {CVPR},
author = {Jen-Hao Rick Chang and Wei-Yu Chen and Anurag Ranjan and Kwang Moo Yi and Oncel Tuzel},
year = {2023},
}

@inproceedings{dai2020neural,
  title={Neural Point Cloud Rendering via Multi-Plane Projection},
  author={Dai, Peng and Zhang, Yinda and Li, Zhuwen and Liu, Shuaicheng and Zeng, Bing},
  booktitle={CVPR},
  pages={7830--7839},
  year={2020}
}

@inproceedings{rublee_orb_2011,
  title={ORB: An Efficient Alternative to SIFT or SURF},
  author={Rublee, Ethan and Rabaud, Vincent and Konolige, Kurt and Bradski, Gary},
  booktitle={ICCV},
  pages={2564--2571},
  year={2011}
}

@InProceedings{Mu_2025_ICCV,
  title     = {Diff2I2P: Differentiable Image-to-Point Cloud Registration with Diffusion Prior},
  author    = {Mu, Juncheng and Ren, Chengwei and Zhang, Weixiang and Pan, Liang and Zhang, Xiao-Ping and Gao, Yue},
  booktitle = {ICCV},
  year      = {2025},
  pages={25777--25787}
}

@article{michiels_tracking_2024,
  title={Tracking and co-location of global point clouds for large-area indoor environments},
  author={Michiels, Nick and Jorissen, Lode and Put, Jeroen and Liesenborgs, Jori and Vandebroeck, Isjtar and Joris, Eric and Van Reeth, Frank},
  journal={Virtual Reality},
  volume={28},
  pages={106},
  issue={2},
  year={2024}
}

@article{SCHUETZ-2021-PCC,
  title={Rendering Point Clouds with Compute Shaders and Vertex Order Optimization},
  author={Sch{\"u}tz, Markus and Kerbl, Bernhard and Wimmer, Michael},
  journal={Computer Graphics Forum},
  volume={40},
  number={4},
  pages={115--126},
  year={2021}
}

@misc{grupp2017evo,
  title={evo: Python package for the evaluation of odometry and SLAM.},
  author={Grupp, Michael},
  howpublished={\url{https://github.com/MichaelGrupp/evo}},
  year={2017}
}

@article{campos_orb-slam3_2021,
  title={ORB-SLAM3: An Accurate Open-Source Library for Visual, Visual-Inertial and Multi-Map SLAM},
  author={Campos, Carlos and Elvira, Richard and Rodríguez, Juan J. Gómez and Montiel, José M. M. and Tardós, Juan D.},
  journal={IEEE Transactions on Robotics},
  volume={37},
  number={6},
  pages={1874--1890},
  year={2021}
}

@article{wisth_vilens_2023,
  title={VILENS: Visual, Inertial, Lidar, and Leg Odometry for All-Terrain Legged Robots},
  author={Wisth, David and Camurri, Marco and Fallon, Maurice},
  journal={IEEE Transactions on Robotics},
  volume={39},
  number={1},
  pages={309--326},
  year={2023}
}

@article{cattaneo_cmrnext_2025,
    title={CMRNext: Camera to LiDAR Matching in the Wild for Localization and Extrinsic Calibration}, 
    author={Daniele Cattaneo and Abhinav Valada},
    journal={IEEE Transactions on Robotics}, 
    year={2025},
    volume={41},
    pages={1995--2013}
}

@article{tao_oxford_2025,
title={The Oxford Spires Dataset: Benchmarking Large-Scale LiDAR-Visual Localisation, Reconstruction and Radiance Field Methods},
author={Tao, Yifu and Mu{\~n}oz-Ba{\~n}{\'o}n, Miguel {\'A}ngel and Zhang, Lintong and Wang, Jiahao and Fu, Lanke Frank Tarimo and Fallon, Maurice},
journal={International Journal of Robotics Research}, 
year={2025},
volume={45},
issue={6},
pages={839--857}
}

@inproceedings{li_deepi2p_2021,
  title={DeepI2P: Image-to-Point Cloud Registration via Deep Classification},
  author={Li, Jiaxin and Hee Lee, Gim},
  booktitle={CVPR},
  pages={15955--15964},
  year={2021}
}

@inproceedings{arandjelovic_netvlad_nodate,
  title={NetVLAD: CNN Architecture for Weakly Supervised Place Recognition},
  author={Arandjelovi{\'c}, R. and Gronat, P. and Torii, A. and Pajdla, T. and Sivic, J.},
  booktitle={CVPR},
  year={2016}
}

@inproceedings{berton_eigenplaces_2023,
  title={EigenPlaces: Training Viewpoint Robust Models for Visual Place Recognition},
  author={Berton, Gabriele and Trivigno, Gabriele and Caputo, Barbara and Masone, Carlo},
  booktitle={ICCV},
  pages={11046--11056},
  year={2023}
}

@inproceedings{berton_megaloc_nodate,
  title={MegaLoc: One Retrieval to Place Them All},
  author={Berton, Gabriele and Masone, Carlo},
  booktitle={CVPR Workshops},
  pages={2886--2892},
  year={2025}
}

@inproceedings{sarlin2019coarse,
  title     = {From Coarse to Fine: Robust Hierarchical Localization at Large Scale},
  author    = {Paul-Edouard Sarlin and
               Cesar Cadena and
               Roland Siegwart and
               Marcin Dymczyk},
  booktitle = {CVPR},
  year      = {2019},
  pages={12716--12725}
}

@inproceedings{hyeon2021pose,
  title={Pose Correction for Highly Accurate Visual Localization in Large-Scale Indoor Spaces},
  author={Hyeon, Janghun and Kim, Joohyung and Doh, Nakju},
  booktitle={ICCV},
  pages={15974--15983},
  year={2021}
}

@article{zhang_visual_2026,
	title = {Visual localization in {3D} maps: comparing point cloud, mesh, and {NeRF} representations},
	volume = {50},
	issn = {1573-7527},
	number = {1},
	journal = {Autonomous Robots},
	author = {Zhang, Lintong and Tao, Yifu and Lin, Jiarong and Zhang, Fu and Fallon, Maurice},
	month = mar,
	year = {2026},
	pages = {14},
}
